\documentclass[10pt, a4paper]{elsarticle} 

\usepackage{float}
\usepackage{amsmath}
\usepackage{graphicx}
\usepackage[unicode=true, bookmarks=false]{hyperref}
\usepackage{times} 
\usepackage[verbose,tmargin=1in,bmargin=1in,lmargin=1in,rmargin=1in]{geometry}
\usepackage{setspace}
\usepackage{mathrsfs}%
\usepackage{appendix}%
\usepackage[x11names]{xcolor}%
\usepackage[]{xcolor}%
\usepackage{textcomp}%
\usepackage{manyfoot}%
\usepackage{booktabs}%
\usepackage{algorithm}%
\usepackage{algorithmicx}%
\usepackage{algpseudocode}%
\usepackage{listings}%
\usepackage{subcaption}
\usepackage{tabularx}
\usepackage{natbib}

\usepackage[superscript]{cite}
\usepackage{xcolor}
\usepackage{titlesec}
\titleformat*{\section}{\large\bfseries\sffamily}
\titleformat*{\subsection}{\normalsize\bfseries\sffamily}
\titleformat*{\subsubsection}{\small\bfseries\sffamily}
\newcommand{\absdiv}[1]{
  \par\addvspace{.5\baselineskip}
  \noindent\textbf{#1}\quad\ignorespaces}
\renewenvironment{abstract}{
    \global\setbox\absbox=\vbox\bgroup
  \hsize=\textwidth
  \noindent\unskip\textbf{\large Summary}
  \par\medskip\noindent\unskip\ignorespaces
  }{\egroup}
\makeatletter
\renewcommand\@biblabel[1]{#1} 
\makeatother

\begin{document}
\begin{frontmatter}

\title{
The Potential of LLMs in Medical Education: Generating Questions and Answers for Qualification Exams
}

\author{
  Yunqi Zhu\textsuperscript{a,d,1}, \sep
  Wen Tang\textsuperscript{b,1}, \sep
  Huayu Yang\textsuperscript{b}, \sep
  Jinghao Niu\textsuperscript{a}, \sep
  Liyang Dou\textsuperscript{b}, \sep
  \\
  Yifan Gu\textsuperscript{c}, \sep
  Yuanyuan Wu\textsuperscript{d}, \sep
  Wensheng Zhang\textsuperscript{a,d,e}, \sep
  Ying Sun\textsuperscript{b,*}, \sep
  Xuebing Yang\textsuperscript{a,*} 
  \\ \vspace{0.5em}
  \emph{\textsuperscript{a}State Key Laboratory of Multimodal Artificial Intelligence Systems, Institute of Automation, Chinese Academy of Sciences, Beijing, China}
  \\
  \emph{\textsuperscript{b}Department of Geriatrics, Beijing Friendship Hospital, Capital Medical University, Beijing, China}
  \\
  \emph{\textsuperscript{c}Xunfei Healthcare Technology Co., Ltd, Beijing, China}
  \\
  \emph{\textsuperscript{d}School of Information and Communication Engineering, Hainan University, Haikou, China}
  \\
  \emph{\textsuperscript{e}Guangzhou University, Guangzhou, China}
  \vspace{-3.5em}
}
















\cortext[corrauth]{Corresponding authors: ysun15@163.com (Ying Sun); yangxuebing2013@ia.ac.cn (Xuebing Yang).}
\fntext[equal]{
    Yunqi Zhu and Wen Tang contributed equally.
}

\begin{abstract}

\absdiv{\textcolor{purple}{Background}}
Large language models (LLMs) have demonstrated successful applicabilities in various domains, from understanding natural language to imitating cognitive inference.
Applications of LLMs on tasks such as the automation of medical report generation and summarization, diagnostic reasoning and personalized treatment recommendations.
Considering most LLMs as answer generator in conversational interactions, whether LLM is capable of being a question setter remains challenging.
In specific to medical scenario, utilizing LLMs to generate questions and answers with accumulated electronic health records (EHRs) will lead to huge potential in revolutionizing the  paradigm of medical eduction.

\absdiv{\textcolor{purple}{Methods}}
In this work, 
we leverage LLMs to produce medical qualification exam questions and the corresponding answers through few-shot prompts, 
investigating in-depth how LLMs meet the requirements in terms of coherence, evidence of statement, factual consistency, and professionalism etc.
Utilizing a multicenter bidirectional anonymized database with respect to comorbid chronic diseases, named Elderly Comorbidity Medical Database (CECMed),
we tasked LLMs with generating open-ended questions and answers based on a subset of sampled admission reports. For CECMed,  
the retrospective cohort includes patients enrolled from January 2010 to January 2022 while the prospective cohort from January 2023 to November 2023, with participants sourced from selected tertiary and community hospitals across the southern, northern, and central regions of China.
A total of 8 widely used LLMs were used, 
including ERNIE 4, ChatGLM 4, Doubao, Hunyuan, Spark 4, Qwen, Llama 3, and Mistral. 
Furthermore, we involved human experts to author the medical examination questions and answers with the same set of reference and prompts, and an independent group of experts to manually evaluate these open-ended questions and answers across multi-dimensional criteria with a 5-point Likert Scale method. 

\absdiv{\textcolor{purple}{Findings}}
For question generation, 
ERNIE 4 achieved the highest cumulative score (16.47).
Human experts scored higher than all AI models in terms of sufficiency of key information (3.67 [3.40, 3.93]), but lagged behind all LLMs in terms of information accuracy (3.63 [3.37, 3.90]).
Notably, ERNIE 4 (3.57 [3.18, 3.96]), Spark 4 (3.57 [3.09, 4.04]), and Mistral (3.53 [3.16, 3.91]) showed a significant advantage over Qwen, in terms of the sufficiency of key information.
For answer generation, 
Human achieved the highest cumulative score (14.50).
Doubao significantly outperformed Mistral in terms of coherence (3.57 [3.25, 3.89]), factual consistency (3.60 [3.30, 3.90]), evidence of statement (3.57 [3.23, 3.90]), and professionalism (3.53 [3.21, 3.85]). Spark 4, demonstrated significant superiority over Mistral on three metrics: coherence (3.53 [3.21, 3.85]), factual consistency (3.57 [3.25, 3.89]), and evidence of statement (3.60 [3.28, 3.92]).

\absdiv{\textcolor{purple}{Interpretations}}
Conventional medical education requires sophisticated clinicians to formulate questions and answers based on prototypes from EHRs, which is heuristic and time-consuming. 
We found that mainstream LLMs could generate questions and answers with real-world EHRs at levels close to clinicians. 
Although current LLMs performed dissatisfactory in some aspects, medical students, interns and residents could reasonably make use of LLMs to facilitate understanding. 


\end{abstract}


\end{frontmatter}

\newpage
\section{Introduction}\label{sec1}

In the interdisciplinary field of artificial intelligence (AI) and medicine, 
the rise of deep learning has brought great opportunities across various clinical applications.  With the booming development of Transformer-based models\cite{vaswani2017attention, Radford2018GPT1, palm2024, touvron2023llama, openai2024gpt4}, AI has fantastic ability to process long contextual semantics for question answering, enabling the generation of coherent and comprehensive output. The success of LLMs stuns medical domain, and the potential of AI in doctor-patient dialogues has attracted great attention for clinicians. Along this path, some researchers have illuminated the convenience of LLMs and remarked some worrying aspects, such as  the persistence of biases in the content generation within the medical domain\cite{ZACK2024e12}, the pronounced reliance on demographic and disease-related decision-making reasoning in LLMs, 
the lack of interpretability and access to private medical data\cite{Yang2024TheLO}, etc. Suggestions have been made to employ data synthesis methods and interactive learning frameworks to elevate the quality of LLMs\cite{ONG2024e428}. In addition, the incorporation of multimodal information in LLMs is advocated to cater to the personalized needs of both patients and healthcare providers\cite{Mes2023npj, Mes2023JMIR}.

\begin{figure*}[htbp]
\centering
\includegraphics[width=1.0\textwidth]{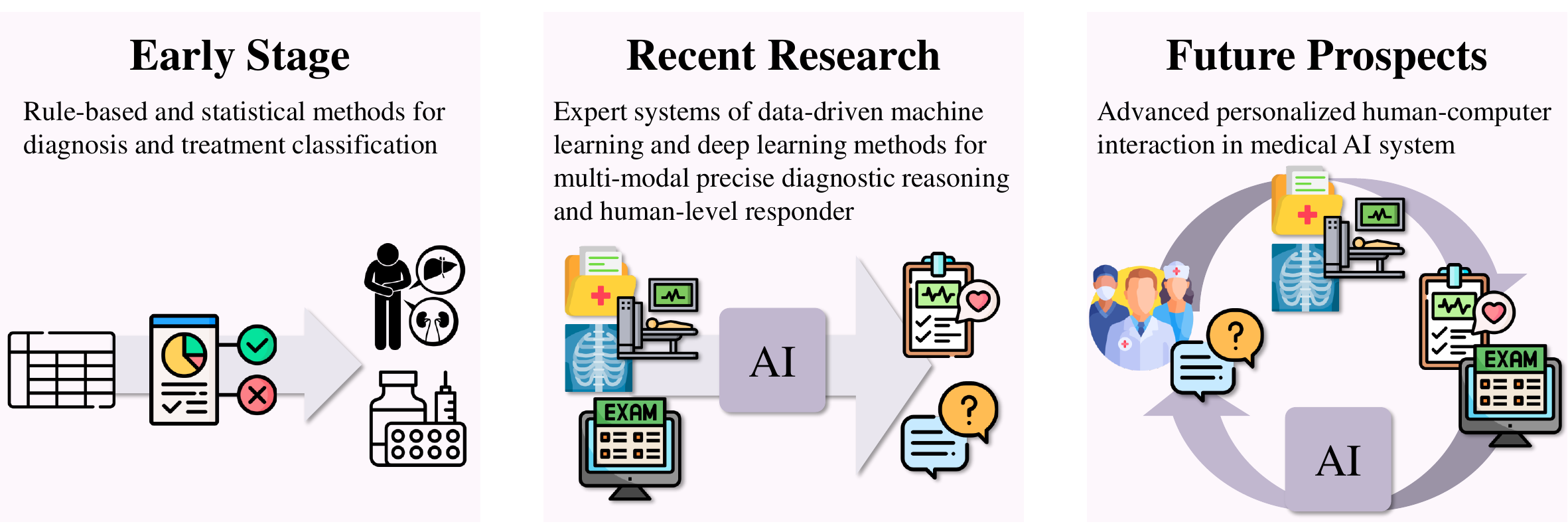}
\caption{
Evolution of the ``AI + Medicine'' paradigm.
}\label{paradigm-fig}
\end{figure*}

Conventionally, AI models in medical education have primarily functioned as responders, tasked with simulating human-like responses in scenarios such as counselor in doctor-patient dialogues\cite{li2021kamed} and examinee in medical exams\cite{AMBOSS, USMLEeasy} Since LLMs are predominantly utilized as answer generators, whether LLMs can serve as a good examiner remains an unresolved challenge. Besides, real-world medical exams require experienced clinicians write questions and answers by drawing upon prototypes from EHRs. This traditional paradigm is characterized by its subjective nature, lack of diversity, and is notably time-intensive. Therefore, this study introduces a novel paradigm where LLMs take on the role of question setters with EHRs, which enables LLMs to create the exam question and answer. This paradigm shift paves the way for personalized learning experiences tailored to the needs of medical students, interns, and residents, ultimately enhancing the effectiveness and efficiency of medical education. Figure~\ref{paradigm-fig} illustrates the paradigm shift of ``AI + Medicine''.

The integration of AI and medicine presents a set of challenges. Universally, concerns regarding the interpretability and reliability of AI-generated outputs, particularly the issue of hallucination\cite{ji2023surveyHall, rawte2023emerHall, Huang2024llmHalluc, zhang2024llmHallSnowball, xu2024HallInevit}, necessitate careful evaluation and mitigation strategies. Ensuring coherence, comprehensiveness, and professionalism is crucial for building trust in AI-generated information. Additionally, addressing biases, interpretability, and privacy concerns are vital for ethical and effective deployment. In the context of medical education, specific challenges arise due to the specialized nature of medical knowledge and the need for personalized learning experiences. Furthermore, examples of human-authored subjective questions and answers are critical for transferring LLMs' knowledge to the specialized medical domain. Meanwhile, there are no standardized benchmarks and evaluation metrics for medical questions and answers. Addressing these challenges requires a multidisciplinary approach, combining expertise in AI and medicine. Collaborative efforts between researchers, clinicians, and educators are essential to develop robust models, establish evaluation standards, and ensure an ethical and effective use of AI in transforming medical education.

\begin{figure*}[htbp]
\centering
\includegraphics[width=0.75\textwidth]{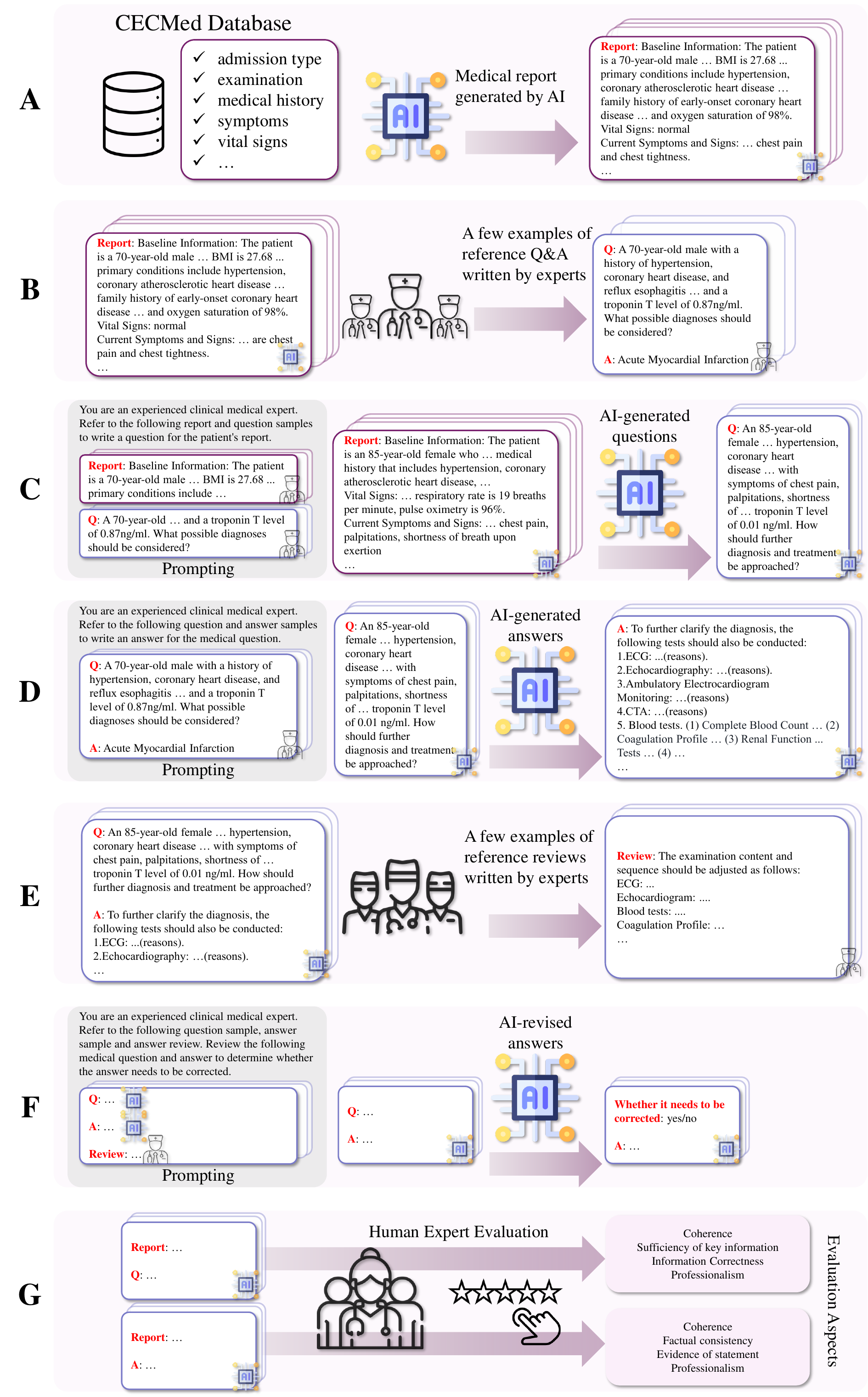}
\caption{
Overall pipeline.
\textbf{A}. Based on a real-world Chinese elderly chronic disease database, 
an AI model was employed to generate patient reports. 
\textbf{B}. Medical experts wrote free-form reference Q\&A for a subset of the reports.
\textbf{C}. The reference report-question pairs were used as prior information, 
combined with prompt templates, 
enabling the AI model to generate questions for the remaining reports.
\textbf{D}. The reference Q\&A pairs were employed as prior information, 
integrated with prompt templates, 
leveraging the AI model to generate answers for the questions produced in phase C. 
\textbf{E}. human experts reviewed and revised a small number of Q\&A pairs.
\textbf{F}. The Q\&A-review pairs were used as prior information, 
integrated with prompt templates, 
leveraging the AI model to decide whether the remaining Q\&A pairs require revision, 
and to provide revised answers if necessary.
\textbf{G}. Based on multiple evaluation aspects, an independent group of human experts assessed the AI-generated Q\&A through scoring.
}\label{pipeline-fig}
\end{figure*}

This absence of consensus on LLMs’ evaluation metrics hinders the ability to compare and assess the performance of different LLMs accurately. The evaluation has been approached with a set of criteria such as coherence, comprehensiveness, professionalism, and readability for diagnostic summary generation\cite{Zhang2024closegapmedSum}. In recent studies, the assessment of AI-generated medical advice has expanded to include empathy, comprehensiveness, and acceptance\cite{Reis2024aidigitalmed}. Moreover, the application of generative AI in the medical field is expected to adhere to principles of accountability, model transparency, content privacy protection, fairness, and harmlessness\cite{Ning2024genAI}. These principles are essential for ensuring the ethical and effective deployment of LLMs in healthcare, thereby overcoming the current limitations and realizing their full potential in transforming medical practice.

To promote the future research and application of LLMs in the medical field, this work concentrates on the task of generating questions and answers for medical qualification exams intended for medical education. Unlike typical medical text generation tasks, such as doctor-patient dialogue, question answering, report summarization etc.\cite{Qiu2023jbhiAI, Singhal2023medpalm, Wang2023HuaTuo, Bolton2024BioMedLM, li2023llavamed}, which evaluate the model's ability to simulate human's responses, this task assesses the model's capability to educate by creating exam content. This work is grounded in a real-world Chinese dataset of elderly chronic diseases, with a primary focus on utilizing the few-shot prompting method for generating both the question and the answer. Specifically, we used a total of 8 LLMs: ERNIE 4, ChatGLM 4, Doubao, Hunyuan, Spark 4, Qwen, Llama 3, and Mistral\cite{baiduai2024,zhipuai2024, volcengine2024, tengcent2024, iFLYTEK2024, aliyun2024, jiang2023mistral7b, grattafiori2024llama3herdmodels}. In addition, we also tasked LLMs with correcting and refining their original AI-generated answers by prompting a few samples of human experts' feedback on the AI-generated answers that have deficiencies.

For comparison, we engaged an independent group of human experts to author medical examination questions and answers based on the same reference cases and prompts.
For evaluation criteria, we refer to the aspects required for recent generative AI in the medical field, such as comprehensiveness, consistency, readability, empathy, equity, non-maleficence, and accountability\cite{Zhang2024closegapmedSum, Reis2024aidigitalmed, Ning2024genAI}. Therefore, we have involved medical experts in the evaluation of the AI-generated content with a 5-point Likert Scale\cite{likert1932A, Joshi_Kale_Chandel_Pal_2015} that a commonly employed scoring method for questionnaires. Specifically, we considered aspects of coherence, sufficiency of key information, information correctness, and professionalism as the evaluation criteria for question generation, and considered aspects of coherence, factual consistency, evidence of statement, and professionalism as the evaluation criteria for answer generation. Figure~\ref{pipeline-fig} visualizes the overall pipeline of this study.

\section{Results}\label{sec2}

\begin{table}[htpb]
\centering
\caption{
  Evaluation results. 
  Question generation is assessed from four aspects: 
  coherence, 
  sufficiency of key information (Suff. key info.), 
  information correctness (Info. corr.), 
  and professionalism (Prof.).
  Answer generation is assesssed from four aspects:
  coherence, 
  factual consistency (Fact. cons.), 
  evidence of statement (Evidence Stat.), 
  and professionalism (Prof.).   
}
\label{result-all-table}

\subfloat[Question generation\label{result-question-generation}]{
\scalebox{1.00}{
\begin{tabular}{lllll}
  \toprule
  & {\:\:\:} Coherence & Suff. key info. & {\:\:\:} Info. corr. & {\:\:\:\:\:\:\:\:} Prof. \\
  \midrule
  Human & 4.13 \footnotesize{[3.94, 4.32]} & 3.67 \footnotesize{[3.40, 3.93]}* & 3.63 \footnotesize{[3.37, 3.90]} & 3.90 \footnotesize{[3.70, 4.10]} \\
  ERNIE 4 & 4.30 \footnotesize{[3.93, 4.67]} & 3.53 \footnotesize{[3.16, 3.91]}* & 4.57 \footnotesize{[4.28, 4.86]} & 4.07 \footnotesize{[3.68, 4.46]} \\
  ChatGLM 4 & 4.17 \footnotesize{[3.73, 4.61]} & 3.37 \footnotesize{[2.93, 3.80]} & 4.43 \footnotesize{[4.18, 4.69]} & 3.70 \footnotesize{[3.24, 4.16]} \\
  Doubao & 4.13 \footnotesize{[3.65, 4.62]} & 3.20 \footnotesize{[2.71, 3.69]} & 4.03 \footnotesize{[3.56, 4.51]} & 3.90 \footnotesize{[3.46, 4.34]} \\
  Hunyuan & 4.30 \footnotesize{[3.87, 4.73]} & 3.33 \footnotesize{[2.83, 3.84]} & 4.20 \footnotesize{[3.73, 4.67]} & 4.07 \footnotesize{[3.56, 4.58]} \\
  Llama 3 70B & 4.10 \footnotesize{[3.68, 4.52]} & 3.47 \footnotesize{[3.04, 3.89]} & 4.43 \footnotesize{[4.06, 4.81]} & 3.93 \footnotesize{[3.47, 4.39]} \\
  Mistral 7B & 4.43 \footnotesize{[4.08, 4.78]} & 3.57 \footnotesize{[3.18, 3.96]}* & 4.43 \footnotesize{[4.10, 4.77]} & 3.90 \footnotesize{[3.38, 4.42]} \\
  Qwen & 3.83 \footnotesize{[3.24, 4.43]} & 2.90 \footnotesize{[2.43, 3.37]} & 4.07 \footnotesize{[3.53, 4.60]} & 3.50 \footnotesize{[2.92, 4.08]} \\
  Spark 4 & 3.80 \footnotesize{[3.22, 4.38]} & 3.57 \footnotesize{[3.09, 4.04]}* & 4.23 \footnotesize{[3.74, 4.73]} & 3.70 \footnotesize{[3.12, 4.28]} \\
  \bottomrule
\end{tabular}
}}

\vspace{1em}

\subfloat[Answer generation\label{result-answer-generation}]{
\scalebox{1.00}{
\begin{tabular}{lllll}
  \toprule
    & {\:\:\:} Coherence & {\:\:\:} Fact. cons. & Evidence Stat. & {\:\:\:\:\:\:\:\:} Prof. \\
  \midrule
  Human & 3.93 \footnotesize{[3.77, 4.10]}* & 3.60 \footnotesize{[3.39, 3.81]}* & 3.53 \footnotesize{[3.32, 3.75]} & 3.43 \footnotesize{[3.18, 3.69]}* \\
  ERNIE 4 & 3.47 \footnotesize{[3.16, 3.77]} & 3.43 \footnotesize{[3.08, 3.78]} & 3.43 \footnotesize{[3.11, 3.75]} & 3.37 \footnotesize{[3.05, 3.68]}* \\
  ChatGLM 4 & 3.23 \footnotesize{[2.82, 3.65]} & 3.33 \footnotesize{[2.88, 3.79]} & 3.30 \footnotesize{[2.84, 3.76]} & 3.20 \footnotesize{[2.75, 3.65]} \\
  Doubao & 3.57 \footnotesize{[3.25, 3.89]}* & 3.60 \footnotesize{[3.30, 3.90]}* & 3.57 \footnotesize{[3.23, 3.90]}* & 3.53 \footnotesize{[3.21, 3.85]}* \\
  Hunyuan & 3.33 \footnotesize{[3.05, 3.62]} & 3.50 \footnotesize{[3.15, 3.85]} & 3.43 \footnotesize{[3.08, 3.78]} & 3.27 \footnotesize{[2.90, 3.63]} \\
  Llama 3 70B & 3.13 \footnotesize{[2.76, 3.51]} & 3.10 \footnotesize{[2.72, 3.48]} & 3.17 \footnotesize{[2.80, 3.53]} & 3.03 \footnotesize{[2.65, 3.42]} \\
  Mistral 7B & 3.07 \footnotesize{[2.70, 3.43]} & 3.07 \footnotesize{[2.69, 3.45]} & 3.07 \footnotesize{[2.69, 3.45]} & 2.93 \footnotesize{[2.58, 3.29]} \\
  Qwen & 3.40 \footnotesize{[3.04, 3.76]} & 3.43 \footnotesize{[3.04, 3.82]} & 3.40 \footnotesize{[3.03, 3.77]} & 3.47 \footnotesize{[3.08, 3.86]}* \\
  Spark 4 & 3.53 \footnotesize{[3.21, 3.85]}* & 3.57 \footnotesize{[3.25, 3.89]}* & 3.60 \footnotesize{[3.28, 3.92]}* & 3.33 \footnotesize{[2.95, 3.72]} \\
  \bottomrule
\end{tabular}
}}
\end{table}


Table~\ref{result-all-table} details the mean scores and their corresponding 95\% confidence interval of the evaluation results.
For question generation, 
ERNIE 4 achieved the highest cumulative score in four evaluation metrics, with a total of 16.47.
Qwen exhibited the overall inferior performance among 8 LLMs.
Further, human experts achieved a total of 15.33.
we consider the result of different methods to be significantly better than Qwen's when the p-value of the Wilcoxon signed-rank test is less than 0.05 (marked with *).
In particular, Human experts scored higher than all AI models in terms of sufficiency of key information (3.67 [3.40, 3.93]), but lagged behind all AI models in terms of information accuracy (3.63 [3.37, 3.90]).
In terms of the internal comparison among LLMs, ERNIE 4 (3.57 [3.18, 3.96]), Spark 4 (3.57 [3.09, 4.04]), and Mistral (3.53 [3.16, 3.91]) secured the top three positions, respectively, in terms of the sufficiency of key information, each demonstrating a significant advantage over Qwen.

For answer generation, 
Human experts achieved the highest cumulative score across four evaluation metrics, totaling 14.50,
while Doubao achieved the second, totaling 14.27.
Mistral showed the overall least favorable performance among 8 LLMs.
we consider the result of LLMs to be significantly better than Mistral's when the p-value of the Wilcoxon signed-rank test is less than 0.05 (marked with *).
It was observed that human experts achieved highest scores on coherence (3.93 [3.77, 4.10]) and factual consistency (3.60 [3.39, 3.81]).
Furthermore, Doubao significantly outperformed Mistral in terms of coherence (3.57 [3.25, 3.89]), factual consistency (3.60 [3.30, 3.90]), evidence of statement (3.57 [3.23, 3.90]), and professionalism (3.53 [3.21, 3.85]). Spark 4, on the other hand, demonstrated significant superiority over Mistral on three metrics: coherence (3.53 [3.21, 3.85]), factual consistency (3.57 [3.25, 3.89]), and evidence of statement (3.60 [3.28, 3.92]). 
Furthermore, ERNIE 4 was found to be significantly more proficient than Mistral in the realm of professionalism (3.37 [3.05, 3.68]).

\begin{figure*}[htbp]
\centering
\includegraphics[width=0.95\textwidth]{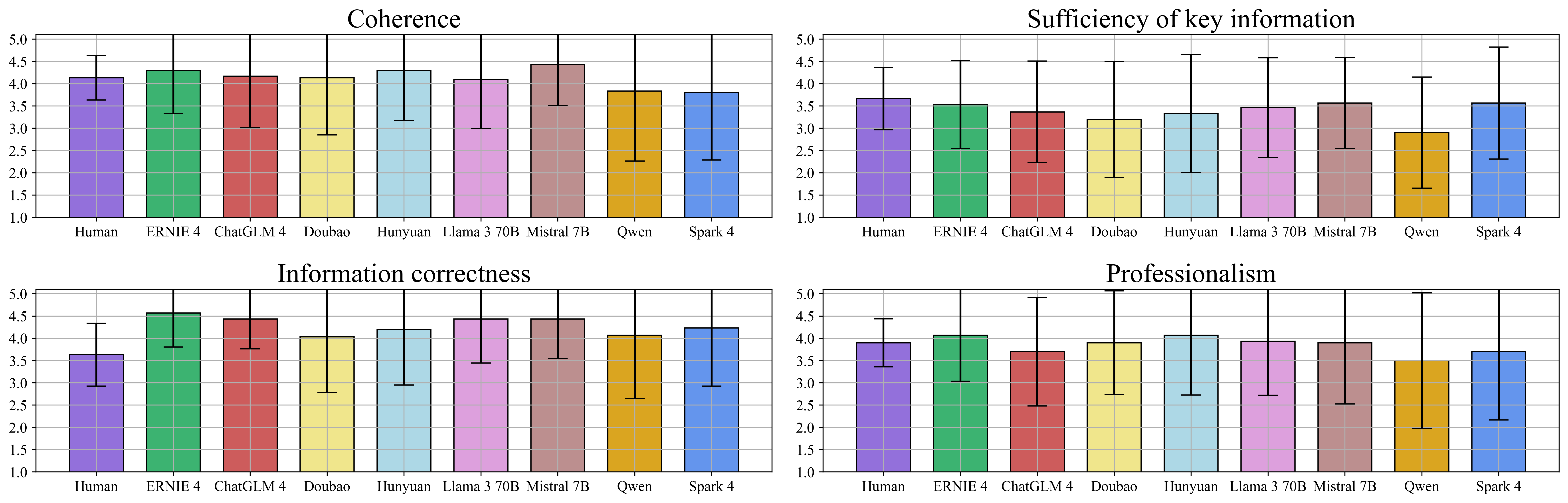}
\vspace{-1.0em}
\caption{
Human evaluation on question generation. Each LLM generated questions based on the same sampled set of admission reports, and each human expert provided scores for all criteria, with the scoring scale ranging from integers 1 to 5, wherein higher scores denoted better performance. Error bars depict the standard deviation of the mean scores.
}\label{Bar-all-question}
\end{figure*}

We implemented a comprehensive qualitative evaluation of the AI-generated questions by human experts, 
focusing on four criteria: coherence, sufficiency of key information, information correctness, and professionalism. 
Figure~\ref{Bar-all-question} visualizes the average rating results of human experts on the question generation of different methods based on a same sampled set of elderly chronic disease admission reports. 
It can be noted that the majority of LLMs achieve scores over 4 in terms of coherence and information correctness, 
and achieve scores nearly 4 in terms of professionalism, whereas the average score for sufficient of key information is relatively lower, with one scoring below 3.5 in general. 
This phenomenon may indicate that human evaluators were satisfied with the contextual semantic correctness and readability of AI-generated questions. 
However, in determining whether the questions are professionally appropriate, 
the LLMs still incur a loss of critical information during the process of extracting and abstracting information from the input reports.
Although the standard deviations of 8 LLMs were greater than the human's, they have achieved a similar criteria level.

\begin{figure*}[t]
\centering
\includegraphics[width=0.95\textwidth]{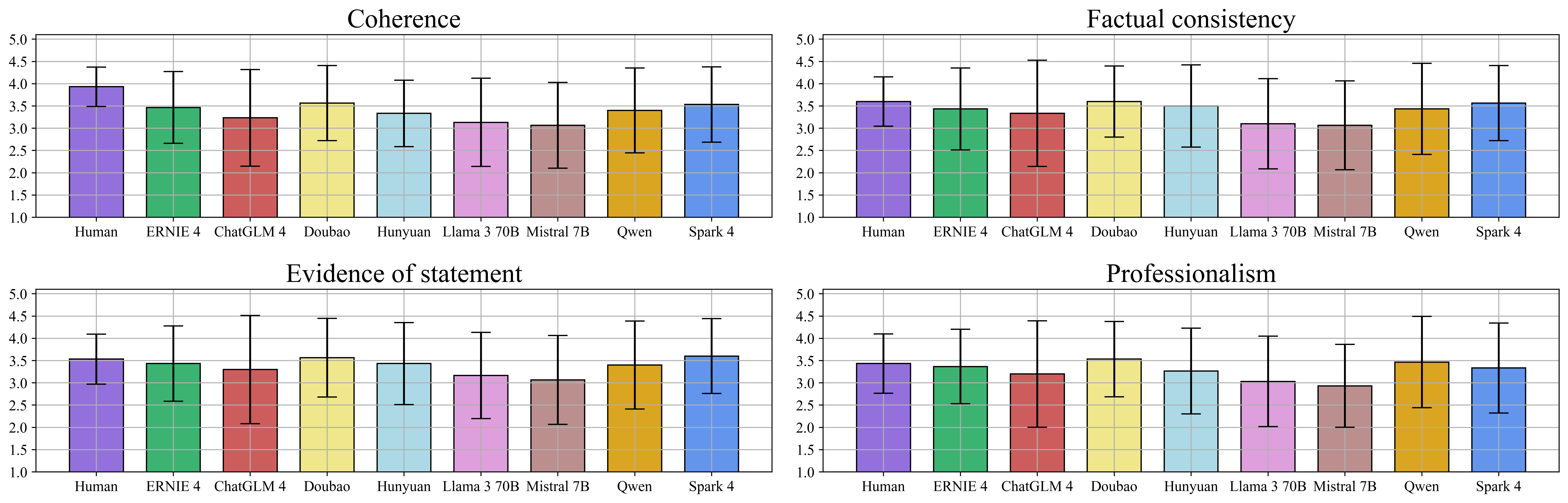}
\vspace{-1.0em}
\caption{
Human evaluation on answer generation. Each LLM generated answers based on the same sampled set of AI-generated questions, and each human expert provided scores for all criteria, with the scoring scale ranging from integers 1 to 5, wherein higher scores denoted better performance. Error bars depict the standard deviation of the mean scores.
}\label{Bar-all-answer}
\end{figure*}

Furthermore, human experts evaluated the AI-generated answers based on four criteria: coherence, factual consistency, evidence of statement, and professionalism. 
Figure~\ref{Bar-all-answer} depicts the average scores of different methods' answers toward a same sampled set of AI-generated questions. 
It can be observed that the average score for question answering is lower than that for question generation, 
with LLMs' ratings hovering around 3.5 across all evaluation metrics. 
Additionally, the standard deviations of human's were generally lower than AI models'.
This indicates a significant gap between the performance of LLMs and the critical requirements of human experts for medical open-ended question answering. 
Moreover, with the task of question answering in the specific domain of elderly chronic diseases under prompting of limited references, 
identifying and improving strategies to refine these four evaluation aspects shows a proper direction for subsequent research efforts.

\begin{figure*}[htbp]
\centering
\includegraphics[width=0.9\textwidth]{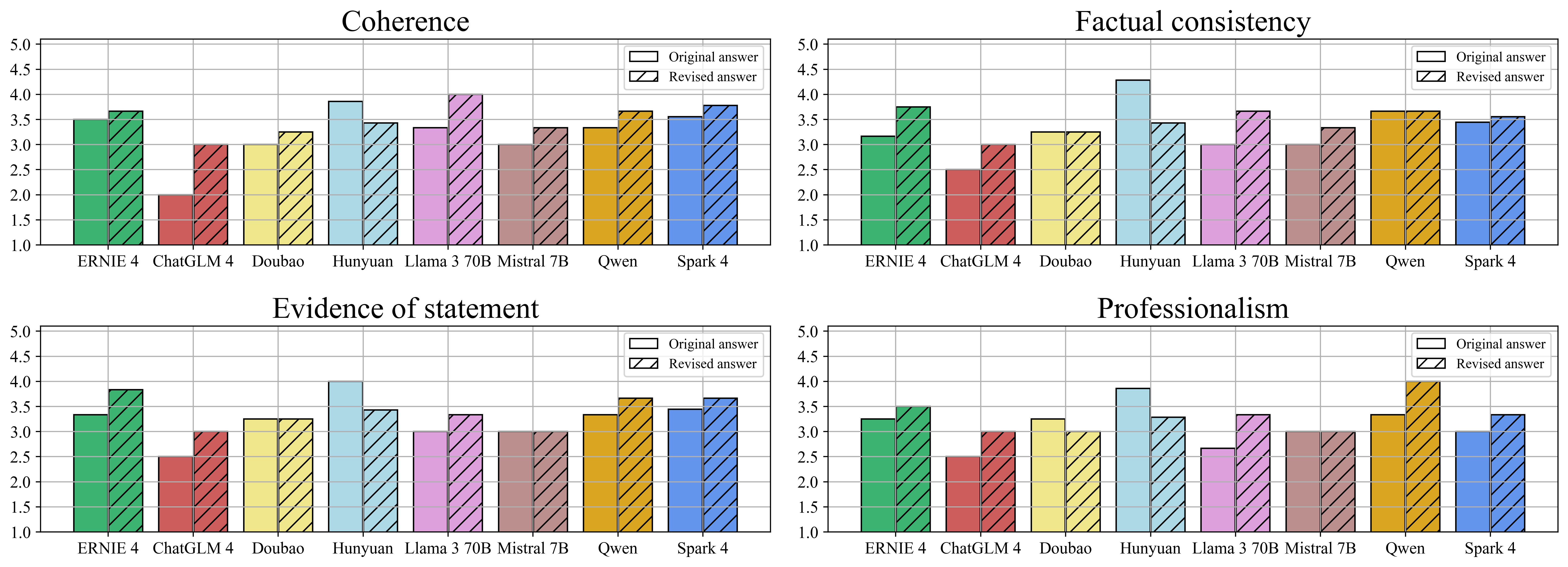}
\vspace{-1.0em}
\caption{
Human evaluation on answer revision. In the histogram, the original answers are unfilled, while the answers based on AI revision are filled with diagonal stripes. Each LLM generated answers based on the same sampled set of AI-generated questions, and each human expert provided scores for all criteria, with the scoring scale ranging from integers 1 to 5, wherein higher scores denoted better performance. Error bars depict the standard deviation of the mean scores.
}\label{Bar-all-revise}
\end{figure*}

Subsequently, we tasked medical experts with evaluating and correcting a subset of flawed AI-generated answers in the form of open-ended short text. We sampled these evaluations and corrections as new prompt engineering materials, 
allowing the model to judge whether a correction is needed for the sampled AI-generated question-answer pairs, 
and to provide new answers. 
We intermingled the answers revised by the model with those in the regular AI-generated answer in the evaluation phase, 
without the human experts being aware of which were directly generated, and which were revised. 
Figure~\ref{Bar-all-revise} shows the comparison of directly generated answers and those modified by the LLMs. 
It can be observed that, across four evaluation metrics, 
at least five models provide correction results that were more satisfactory to the human experts, 
with a few models not showing a change in scores, while the performance of the Hunyuan model slightly deteriorated.

In the supplementary material, we visualize the distributional relationship between the AI-generated questions and answers under different scores and the length of the generated text after the removal of stop words Figure~.1 (a) and (b). 
We also visualize the distributional relationship between the proportion of words in the output text that do not appear in the input text under different scores Figure~.1 (c) and (d). 
It can be observed that, 
except for a slight positive correlation between ``sufficiency of key information'' and text length, 
there is no obviously simple linear relationship between other evaluation metrics and text length or the proportion of novel words. 
Moreover, the average text length for question generation and answer generation are 40.7, and 138.5, respectively. 
And then the proportion of novel words for question generation and answer generation are 7.4\% and 68.8\%, respectively.

\section{Discussion}\label{sec12}

This work introduce the novel paradigm of AI for medical education,
and demonstrates the feasibility of LLMs in generating medical qualification exam questions and answers pertaining to elderly chronic diseases through few-shot prompting, 
with the goal of simulating the question bank of USMLEasy\cite{USMLEeasy} and AMBOSS Qbank\cite{AMBOSS}. The ability of LLMs to simulate real-world medical questions is encouraging, 
as it opens avenues for automated content generation in medical education. 
Nevertheless, a notable discrepancy in the quality of answers underscores the challenges of generating accurate, evidence-based, and professional responses. 
This emphasizes the necessity for further research and development to refine LLMs' comprehension and application of medical knowledge.

One potential direction is to explore the incorporation of medical knowledge base and expert feedback into LLMs using Retrieval-Augmented Generation. 
By integrating structured medical knowledge and leveraging human corrections, 
LLMs can further improve their factual consistency and evidence-based reasoning capabilities. 
Furthermore, employing strategies such as human preference alignment and curriculum learning can help LLMs learn from their mistakes and progressively improve their performance in generating high-quality, in-depth and comprehensive medical responses.

In the supplementary material, we show the specific prompt engineering templates (Table~.1), the reference question and answer pairs (Table~.2), and examples of AI-generated questions and answers (Table~.3 .4 .5 .6).
It can be noted that the AI-generated questions can closely emulate the text style of the reference questions to a significant extent. 
Although the reference answers are concise and very short in length, 
it is likely a consequence of the LLMs' training on extensive dataset for general text generation tasks as well as medical question-answering, 
resulting in the generated answers containing substantial discussion and explanations of causes. 
Moreover, it can be observed that the LLMs, primarily trained on English corpora, 
occasionally produce answers in English (e.g. Llama 3's response in the supplementary material Table~.6 ). 
In addition, AI-generated questions that receive a overall low score may be attributed to the LLMs could omit several medical procedures or tests,
thereby directly inquiring about advanced stages of the patient's treatment. This is a typical case of LLMs' hallucination issue.
For instance, Hunyuan's question in supplementary material Table~.4 should not ask about ``treatment'' (the provided information does not substantiate a diagnosis for the patient, thus bypassing the diagnostic reasoning stage), 
but rather should focus on ``what further tests should be conducted next''.

It is crucial to acknowledge the limitations of the LLMs. 
State-to-the-art generative language models may still produce inaccurate or contradictory information up to now, 
as they are predominantly reliant on the extensive corpus and the probability distributions of the upcoming words and sentences during the autoregressive generation. 
Moreover, the prompt engineering method can stimulate LLMs' pre-trained knowledge within the medical domain and adapt text styles to resemble the reference, 
but it cannot guarantee the qualities of outputs once and for all. 
Consequently, the limited real-world understanding and reasoning abilities of LLMs can potentially result in false conclusions. 
Additionally, the performance of LLMs can be influenced by biases present in the reference materials and the prompt templates, potentially producing discriminatory outputs. 
Efforts are needed to mitigate these biases through subsequent human supervision for the AI-generated content intended for medical education.

In this work, we evaluated the capability of LLMs in generating medical examination questions and answers using a real-world Chinese elderly chronic disease dataset. 
Extending the scope of medical examination to other clinical departments such as oncology, 
gynaecology, and orthopedics could comprehensively assess the performance of LLMs. 
Furthermore, expanding this work to medical examination in other languages or under multilingual settings requires investigation in future studies. 
Additionally, this study examined eight LLMs, predominantly closed-source,
and given that commercial AI companies may update the APIs of these LLMs, 
reproducibility emerges as a significant challenge in evaluating the current performance of the state-of-the-art LLMs.
Additionally, we involved an independent group of human experts to write the questions and answers based on the same reference cases and prompts as those used by the LLMs.

In this study, the evaluation of medical qualification exams was contingent upon human evaluation. 
Due to the resource-intensive and time-consuming nature of having medical experts write comprehensive and open-ended reviews, 
a scoring system was adopted. 
Besides, since this study employed advanced and competitive LLMs for content generation, 
we did not employ additional AI models for assessments such as coherence and counterfactual evaluation which might be ambiguous and foregone. 
Moreover, automated evaluation methods at the level of tokens, words, or phrases are dependent on the reference, 
and the cost associated with eliciting open-ended references from medical experts is prohibitively high. 
These factors highlight the constraints inherent in the evaluation for the task of generating medical qualification exams, 
whereas also points to a potential avenue for subsequent research to investigate.

In summary, 
this study pioneers a novel AI role in medical education by transforming AI from dialogue participant to sophisticated question setter, 
dramatically advancing conventional paradigms and revolutionizing medical education with personalized healthcare insights. 
Concretely, this study provides valuable insights into the application of LLMs for the generation of medical qualification exam questions and answers. 
LLMs showed competitive performance parity with human experts on some evaluation aspects.
Further, although LLMs showed promise in generating questions, 
there remains substantial scope for advancement in the quality of the answers. 
By harnessing the feedback of medical experts using few-shot prompt engineering, 
we can refine LLMs and enhance their potential to contribute to the domain of medical education.

\section{Methods}

\subsection{Data collection}

We established a multicenter bidirectional anonymized database of older patients with comorbid chronic diseases [China Elderly Comorbidity Medical Database (CECMed)]. 
The retrospective cohort had enrollment from January 2010 to January 2022, 
while the prospective cohort had enrollment from January 2023 to November 2023. 
The patients were recruited from selected tertiary hospitals and community hospitals in southern, northern, and central regions of China. 
Inclusion criteria are as follows: 
(1) aged $\geq$ 65 years; 
(2) with at least one of the following five chronic diseases: coronary heart disease (CHD), hypertension, diabetes, chronic obstructive pulmonary disease (COPD) and osteoporosis. 
Exclusion criteria are as follows: 
(1) Late stage malignant tumors, expected survival time less than 3 months; 
(2) Completely disabled and unable to communicate; 
(3) Unable to cooperate with follow-up.

For hospitalized patients, 
administrative information includes admission time, admission type, demographics, socioeconomic, past medical history, previous medication use, current symptoms, vital signs, laboratory tests and examinations, reasons for admission and admission diagnosis. 
All patients underwent comprehensive geriatric assessment within 24 hours after admission. 
Barthel index, FRAIL, MORSE, MNA-SF and MINI-cog were used to assess patient's functional status. 
For outpatient patients, baseline data includes demographics, socioeconomic, medical history, medications, vital signs, comprehensive geriatric assessment and laboratory tests and examinations within the past three months. 
Patient data gathered was anonymized and preprocessed in a standardized manner. 
Furthermore, guided by semi-structured templates that integrate medical expert knowledge, 
we used LLMs to synthesize the information into admission reports and summaries.

This study was registered (\url{www.clinicaltrials.gov}, NCT06316544) and approved by the ethical review boards at the participating hospitals. All patients provided informed consent.

\subsection{Few-shot prompting}

First, sampling from the Chinese elderly chronic disease database, 
human experts composed 4 questions along with corresponding answers in an open-ended question-and-answer format as the reference. 
Subsequently, we randomly sampled 30 admission reports within the dataset, 
and used the reference question as prompts to guide 8 different LLMs for the task of question generation. 
Next, under the guidance of medical experts, we preprocessed the AI-generated questions, 
taking into account diversity and professionalism. We then randomly sampled 30 different questions, 
using the reference question-answer pairs as prompts, and instructed the LLMs to generate responses to these questions. 
In addition, medical experts provided a total of 20 concise open-ended reviews of the AI-generated answers, 
which included correct answers, incorrect answers and the reasons for errors. 
Further, we constructed prompts based on these answer reviews. 
As a result, we obtained 240 questions, 240 direct answer generations, and 43 AI-corrected answers. 
It should be noted that the reports and generated samples that involved human expert authorship and evaluation, 
as mentioned above, were not included in the final phase of human expert scoring.

Few-shot learning is an efficient strategy that leverages a small amount of annotated data to optimize machine learning models for domain-specific tasks. 
This study adopted the method of prompt engineering, enabling LLMs to emulate a limited number of reference medical qualification exam questions and answers, 
thereby eliciting LLMs' knowledge in the medical specialties and adapting to the question generation as well as question-answering task for the medical education.
Moreover, the specific prompt engineering templates are available in the supplementary material Table~.1.
The original Chinese texts and their corresponding English translations are provided.

\subsection{Human evaluation}

We evaluated the performance of LLMs on producing qualification examinations for medical education through medical expert assessment. 
To measure and compare the practical usability of LLMs, we involved an independent group of human experts to author medical examination questions and answers based on the same reference cases and prompts.
Specifically, we focused on two tasks: generating open-ended questions based on admission reports and generating corresponding open-ended answers to the examination questions. 
(1) For the generation of examination questions, we assessed from four perspectives: coherence, sufficiency of key information, information correctness, and professionalism. 
Coherence refers to the contextual relevance and rationality between the information, viewpoints, and sentences, forming a clear and easily comprehensible text. 
Sufficient key information refers to the ability of LLMs to extract, abstract, and organize important and complete information from the input reports, 
which is adequate for the examinee to formulate a reasonable response based on this background information. 
Information correctness indicates whether the question is reliable and valid, and whether the question contradicts the factual knowledge. 
Professionalism is measured by the accuracy of terminology used, the relevance and consistency of medical knowledge content, 
adherence to professional medical ethical standards, and the depth of medical knowledge assessment; 
(2) For question-answering, the evaluation was based on the following four dimensions: coherence, factual consistency, evidence of statement, and professionalism. 
The dimensions of coherence and professionalism were assessed similarly to those mentioned above. 
Factual consistency evaluated whether the model's responses aligned with medical facts and correctly addressed the question. 
Additionally, the evidence of statement examined whether the model's reasoning and conclusions were supported by its supporting evidence within its response. 
Additionally, during the evaluation phase of AI-generated answers, those AI-revised answers are randomly mixed into them.

\section*{Acknowledgement}
This work was supported the National Key R\&D Program of China (No. 2021ZD0111000), the Excellent Youth Program of State Key Laboratory of Multimodal Artificial Intelligence Systems (No. MAIS2024309) and the National Natural Science Foundation of China (No. 62394330).

\bibliographystyle{elsarticle-num} 
\bibliography{elsevier-template}
 
\end{document}